\documentclass{article}

\usepackage{arxiv}

\usepackage[utf8]{inputenc} 
\usepackage[T1]{fontenc}    
\usepackage{hyperref}       
\usepackage{url}            
\usepackage{booktabs}       
\usepackage{amsfonts}       
\usepackage{nicefrac}       
\usepackage{microtype}      
\usepackage{lipsum}		
\usepackage{graphicx}
\usepackage{natbib}
\usepackage{doi}

\title{Generative Design Ideation: A Natural Language Generation Approach }

\date{March 27, 2022}	

\author{ \href{https://orcid.org/0000-0002-5401-6679}{\includegraphics[scale=0.06]{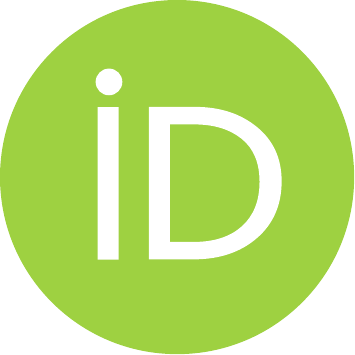}\hspace{1mm}Qihao Zhu}\\
	Data-Driven Innovation Lab\\
	Singapore University of Technology and Design\\
	\texttt{qihao\_zhu@mymail.sutd.edu.sg} \\
	\And
	\href{https://orcid.org/0000-0001-5892-8432}{\includegraphics[scale=0.06]{orcid.pdf}\hspace{1mm}Jianxi Luo} \\
	Data-Driven Innovation Lab\\
	Singapore University of Technology and Design\\
	\texttt{jianxi\_luo@sutd.edu.sg} \\
}

\renewcommand{\shorttitle}{Generative Design Ideation: A Natural Language Generation Approach}

\begin{document}
\maketitle

\begin{abstract}
This paper aims to explore a generative approach for knowledge-based design ideation by applying the latest pre-trained language models in artificial intelligence (AI). Specifically, a method of fine-tuning the generative pre-trained transformer using the USPTO patent database is proposed. The AI-generated ideas are not only in concise and understandable language but also able to synthesize the target design with external knowledge sources with controllable knowledge distance. The method is tested in a case study of rolling toy design and the results show good performance in generating ideas of varied novelty with near-field and far-field source knowledge.
\end{abstract}


\section{Background}
The performance of human design ideation is often limited by the designers’ knowledge scope, which hinders them from exploring external inspiration and opportunities to be applied in the domain of interest. In this section, we discuss the need of generative and knowledge-based AI tools for design ideation process and introduce the recent advance of pre-trained language model that we use in this study to develop such a tool. \par
\label{sec:headings}
\subsection{Computer-Aided Design Ideation}
“What to design?” is a common question for designers and companies when developing innovative products. At the early design stages, the quantity and diversity of conceptual design ideas are essential for designers to explore new design opportunities departing away from existing designs. However, the well-recognized features in existing designs can continuously fixate designers’ thinking and limit their capability of generating novel ideas \citep{viswanathan2016a}. Traditional approaches to overcome design fixation include brainstorming and design heuristics \citep{white2012from,yilmaz2016evidencebased}, which have been proven effective to improve creative thinking and help designers to think out-of-the-box. However, apart from design thinking, the limitation of designers’ knowledge base is also an important source of design fixation.\par
In recent years, knowledge-based methods and tools have been developed for computer-aided ideation. For example, AskNature is a web-based tool that provides a large variety of biomimicry knowledge and strategies for designers to draw inspiration from nature \citep{deldin2014the}. Luo et al introduced InnoGPS to guide the provision of design stimuli from the patent database by their knowledge distance to the design problem or interest \citep{luo2019computer,luo2021guiding}. Sarica et al. developed the technology semantic network for identifying technical white space according to the semantic distance between the design target and the stimuli \citep{sarica2020technet,sarica2021idea}. These approaches are design stimulators that systematically provide knowledge-based inspirational stimuli to provoke designers during ideation.\par
On the other hand, researchers have also been exploring generative approaches that automate the early phase design process. For example, generative grammars are used to encode design knowledge computationally, which can be used to rapidly generate design alternatives \citep{chakrabarti2011computer}. However, the graphical design representation from this approach can be too abstract for designers to articulate, and therefore further inference is needed to generate useful design ideas. Moreover, the recent advance of AI has led to deep generative models, such as Generative Adversarial Networks (GAN) and Variational Autoencoders (VAE), which have been increasingly popular in design automation \citep{regenwetter2021}. These models learn visual design knowledge from images or meshes and generate design concepts also in visual representation. This makes them more suitable for embodiment and detailed design rather than the early-stage design ideation. Therefore, the knowledge-based and generative ideation method is still an open arena in design research.\par
Figure 1 shows different computer-aided ideation modes in terms of human-computer collaboration. Figure 1 (a) is the most common scenario where humans generate ideas, and record, analyze and improve them with the aid of computers. In scenario (b), computers provide stimuli to inspire human designers to generate ideas. Many existing computer tools, such as InnoGPS, can facilitate such a process. In scenario (c), computer-based tools generate the ideas for human designers to evaluate, select and improve. The computer-generated ideas could also provoke designers to be more creative with the extended knowledge that comes with the generated ideas. Scenario (c) is of the central interest of the present study.\par
\begin{figure}[ht]
	\centering
	\includegraphics[width = 15cm]{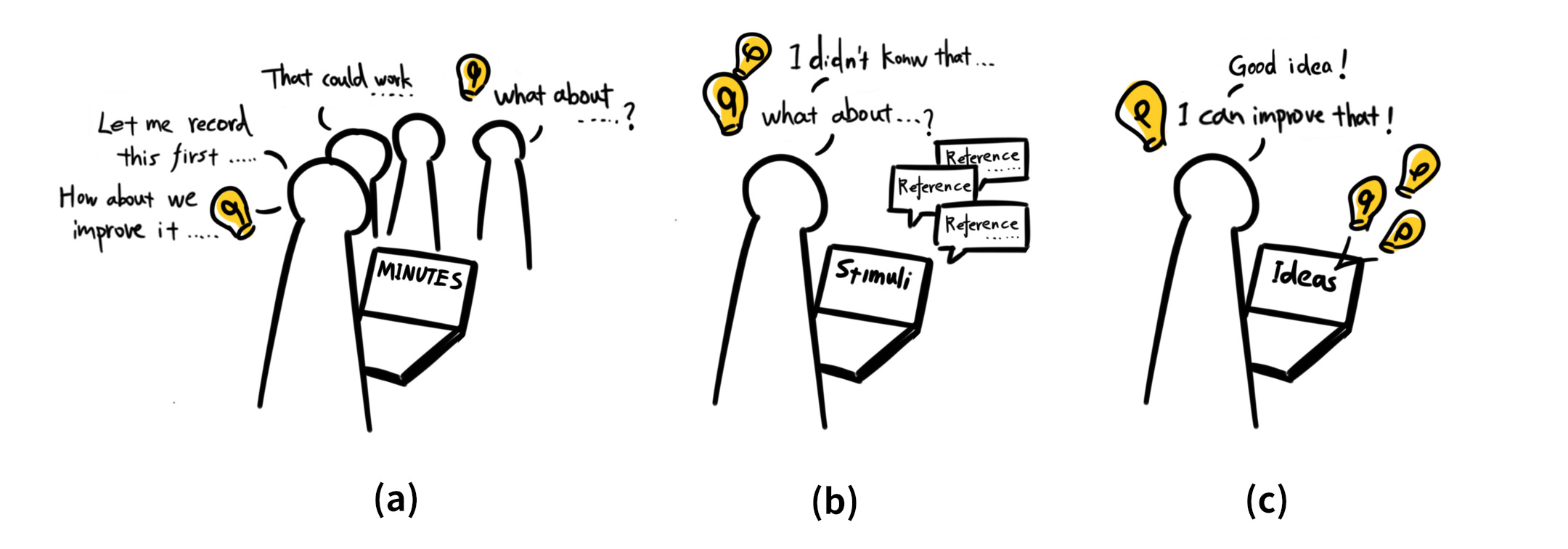}
	\caption{Human-computer collaboration in different ideation approaches: (a) Brainstorming (b) Stimuli-based approach (c) Generative approach}
	\label{fig:fig1}
\end{figure}

\subsection{Pre-Trained Language Model}
Pre-Trained Language Model (PLMs) are language models that have been trained with a large textual dataset collected from varied sources including Wiki, books, and web data, and can be applied to deal with specific language-related tasks \citep{duan2020a}. In recent years, transformer-based language models have been achieving state-of-the-art performance on many tasks. For example, the generative pre-trained transformer (GPT) \citep{radford2019language,brown2020language} is one of the most popular series of PLMs. GPT-2 \citep{radford2019language} is trained on 8 million documents of web data and can perform different downstream language-related tasks after fine-tuning. Fine-tuning is a training technique that uses a relatively small dataset of the task of interest to retrain a large pre-trained model. Since being released, GPT-2 has achieved dramatic improvement on natural language generation (NLG) tasks like natural language inference and question-answering. GPT-3 \citep{brown2020language} is the largest PLM nowadays and was trained on a 500 billion tokens dataset. This powerful model is designed for prompt-based learning that requires only a few examples of the specific task to achieve outstanding performance.\par
Another important transformer-based PLM is the Bidirectional Encoder Representations from Transformers (BERT) \citep{kenton2019bert}, which was trained with Wiki and books data that contains over 3.3 billion tokens. BERT is commonly used in natural language understanding (NLU) tasks such as text classification and keyword extraction. However, as a masked language model, BERT is generally weak at NLG, because it can only learn the contextual representation of words \citep{duan2020a}.\par
The state-of-the-art network architecture of transformers, as well as the huge pre-training datasets, offer PLMs not only the capability of learning human language, but also the knowledge and logic that come with it. By applying PLMs, knowledge-based systems could leverage a wide range of knowledge without the need for manual formulation and inference.\par

\section{Aim}
\label{sec:headings}
Existing computer-aided ideation methods can retrieve external knowledge as stimuli but lack the ability to generate and represent ideas in understandable forms. Therefore, experiences and skills are still required for designers to generate and articulate creative ideas when using these methods, and the quality and quantity of the alternatives for the convergence of ideation are dependent on the designers. Moreover, when using external knowledge as stimuli during ideation, far-field stimuli tend to yield more novel but less feasible ideas comparing to near-field stimuli \citep{luo2019computer,luo2021guiding}. The present research aims to quickly generate ideas in a given domain of interest that take near-field or far-field external knowledge as inspiration, and at the same time, in the understandable representation of natural language.\par
Furthermore, we aim to explore the use of generative pre-trained language models in developing an effective knowledge-based expert system for conceptual design and fill in the gap that transformer-based AI models have yet to be seen utilized in the design community \citep{regenwetter2021}.\par

\section{Significance}
\label{sec:headings}
This research will leverage the recent progress of natural language processing (NLP) to creative design ideation tasks and provide new insights into computer-aided design ideation. For novice designers, engineers, or students, this approach can help them think out of the box. It can also help experienced designers ideate with an extended knowledge base. Moreover, the generated ideas in human language could provoke the designers for even more ideas, like the inspirational human-to-human interactions in the brainstorming process, just this time the brainstorming will be happening between humans and computers (Figure 1. c). This will not only boost the diversity and speed of the generation of initial ideas but also potentially change the human-computer interaction during ideation.\par

\section{Method}
\label{sec:headings}
\citet{luo2019computer} introduced InnoGPS (http://www.innogps.com/), a knowledge-based design expert system based on all USTPO patents from 1974 to 2020 that guides the retrieval of near-domain and far-domain stimuli by knowledge distance to aid creative design ideation. The system uses a network of all international patent classes to store, organize and retrieve the world’s cumulative technical design knowledge according to statistically estimated knowledge distance between patent classes. In InnoGPS 2.0 \citep{luo2021guiding}, the design stimuli are provided at the semantic, document and field levels simultaneously. While providing data-driven ideation aids, InnoGPS itself does not have generative capability. In this work, we employ InnoGPS to gather patent data and assess knowledge distance. \par 
Figure 2 shows the dataset preparation, fine-tuning, and idea generation processes in our method. For a specific domain as the knowledge source, we gather the patent titles in this domain from InnoGPS and extract a keyword for each title using KeyBERT. KeyBERT \citep{sharma2019self} leverages BERT embeddings to extract keywords and key phrases that best represent a document. The prepared dataset containing keyword-title pairs in the selected domain is then used to fine-tune the base model of GPT-2. Finally, given any target keyword representing the home design, the fine-tuned model will generate ideas for the target design based on external knowledge learned from the selected domain. The example in Figure 2 shows how the model can learn from an external domain and generate rolling toy ideas based on it.\par
\begin{figure}
	\centering
	\includegraphics[width = 9cm]{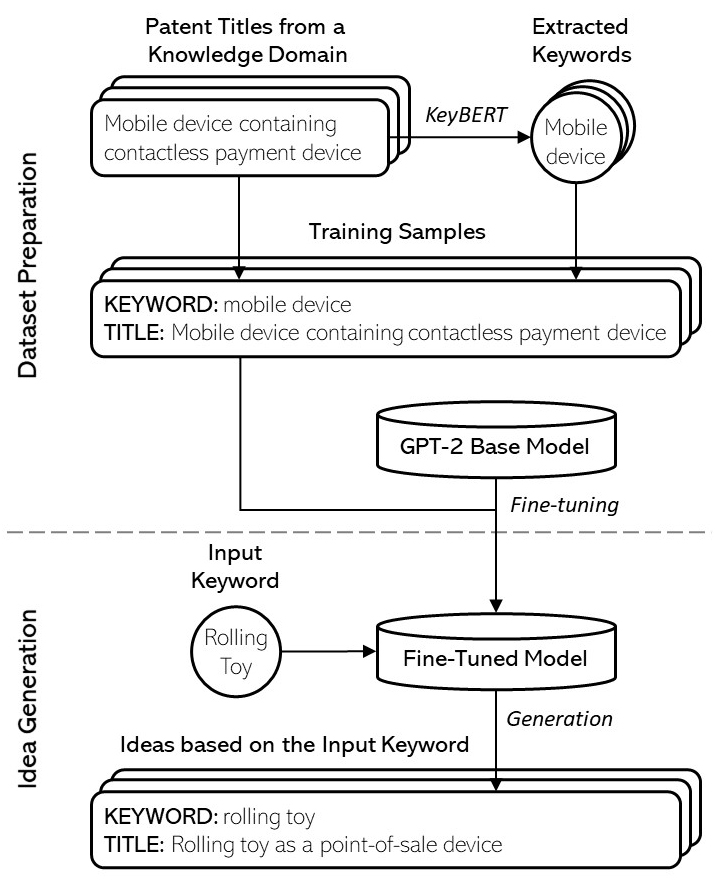}
	\caption{GPT-2-based idea generation workflow}
	\label{fig:fig2}
\end{figure}

In this way, thanks to the knowledge learned from both the commonsense pre-training and domain-specific fine-tuning data, the model can leverage the source knowledge in a way that is potentially applicable to the target, and thus generate novel and useful ideas. \par
In a practical design ideation scenario, the proposed method can be used with InnoGPS to explore external knowledge domains before generating ideas. Figure 3 depicts the workflow. After inputting a keyword of the target domain of interest in InnoGPS, the user will be directed to a list of nearby domains in the order of their knowledge distance proximity (as quantified in InnoGPS) to the target domain. Selecting a domain as knowledge source will generate ideas based on both the target and source domains. Each knowledge domain in the expert system, e.g., InnoGPS in this case, can have a fine-tuned GPT-2 model to generate ideas as the user may request, just like a virtual domain expert that can ideate for the user. Different virtual experts whose knowledge with varied knowledge distance to the target design may help generate new ideas with varied novelty and feasibility.\par
\begin{figure}
	\centering
	\includegraphics[width = 11cm]{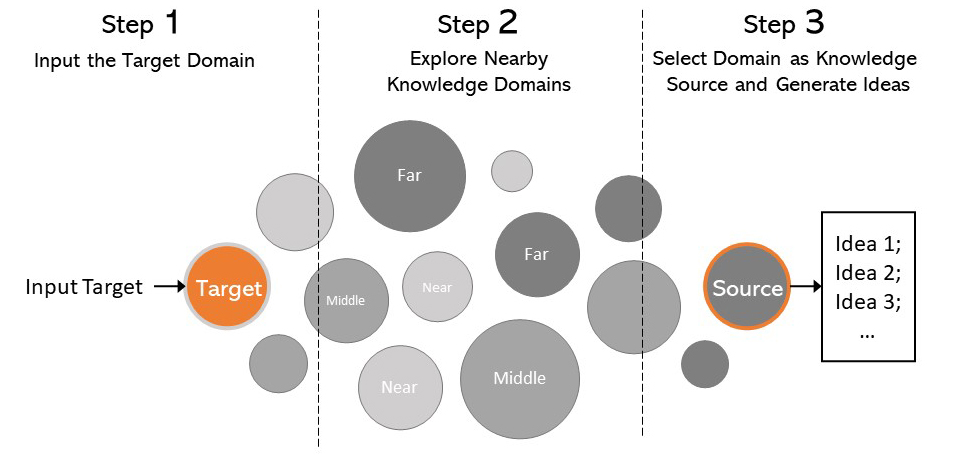}
	\caption{Workflow of using the proposed method together with InnoGPS}
	\label{fig:fig3}
\end{figure}

\section{Case Study and Results}
\label{sec:headings}

Consistent with the case study conducted by \citet{luo2019computer}, this experiment focuses on the design ideation of rolling toys and retrieving design stimuli from near-domain and far-domain knowledge. Six domains are picked as references according to their rank order by knowledge proximity to rolling toys: 1) Weapons, 2) Agriculture, 3) Lighting, 4) Drilling \& Mining, 5) Grinding \& Polishing, and 6) Fuels \& Lubricants. Regarding knowledge distance calculation, please refer to \citet{luo2019computer,luo2021guiding}. The first three domains are relatively near-field (13$^{th}$, 17$^{th}$, 28$^{th}$ nearest, respectively) while the other ones are considered far-field (64$^{th}$, 68$^{th}$, 100$^{th}$ nearest, respectively). These domains vary a lot regarding the number of patents they hold. To control variables and test for performance, we picked the latest 20,000 patents from each domain to form our fine-tuning dataset. Moreover, the titles of the chosen patents all have more than three words because we want the model to learn to generate ideas that contain more information. Otherwise, the model will be more likely to simply repeat the keyword without adding any useful insights. \par
For this study, we use the 355M base model of GPT-2, and each model is fine-tuned for 20,000 steps with a batch size of 1. Figure 4 shows the fine-tuning loss of each model during the first 10,000 steps. The loss is stabled afterward. The fine-tuning loss is plotted every 100 steps.\par

\begin{figure}[ht]
	\centering
	\includegraphics[width = 12cm]{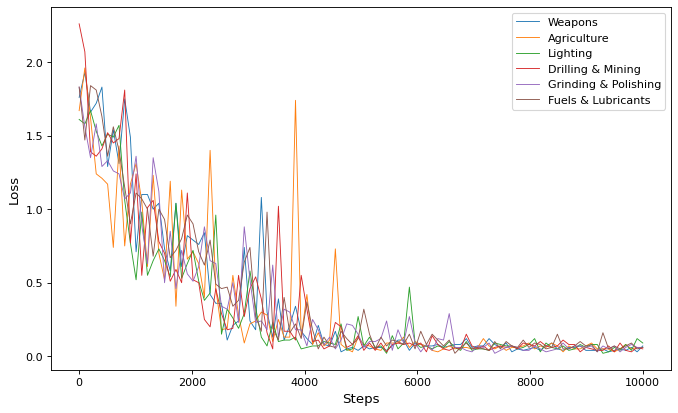}
	\caption{Training loss during the fine-tuning of the models}
	\label{fig:fig4}
\end{figure}
For idea generation, we use the parameters of ‘temperature=0.9; top-k=50’ to enable a more randomized generation for more creative ideas. 500 rolling toy ideas are generated by the virtual domain expert (i.e., fine-tuned GPT-2 based on each domain’s knowledge) and the unique ones are selected for further analysis. Table 1 reports the percentage of unique ideas from each domain and some examples of generated rolling toy ideas. \par
In general, the proposed method can generate understandable and concise ideas in natural language that take advantage of external knowledge. This can be potentially valuable for the ideation with both near-field and far-field knowledge sources. For example, given near-field stimuli of weapons, rolling toy designers may quickly come to the idea of “adding a gun or other shooting mechanism to the rolling toy” and then fixate on it.  Our model, in addition to that, suggests that the rolling toy can also be designed into a moving shooting target or dart board based on the domain of weapons. For far-field stimuli, the model can also provide ideas as inspiration when designers find it challenging to synthesize such knowledge.\par

\begin{table}[h]
	\caption{Results of rolling toy idea generation}
	\centering
	\begin{tabular}{p{3cm}p{2.5cm}p{9cm}}
		\toprule
		Domain     & \% Unique Ideas     & Examples of Generated Ideas\\
		\midrule
		Weapons\par(13$^{th }$ nearest) & 35.8\% (179/500)   & •	Rolling toy wheeled target.\par•	Rolling toy projectile launcher.\par•	Rolling toy dart board capable of making turns.\par•	Rolling toy air gun.
    \\
		\midrule
		Agriculture\par(17$^{th }$ nearest) & 40.8\% (204/500)   & •	Rolling toy bale wrapper apparatus.\par•	Rolling toy saddle with pressure adjustment.\par•	Rolling toy bump stop device.\par•	Rolling toy with liquid container.
\\
		\midrule
	Lighting\par(28$^{th }$ nearest) & 69.6\% (348/500)   & •	Color changing LED roll toy.\par•	Musical spell-playing rolling toy.\par•	Rolling toy and cart with a plurality of removable LED-units.\par•	Lighting device for rolling and adjusting a light source.
\\
		\midrule
		Drilling \& Mining\par(64$^{th }$ nearest) & 69.8\% (349/500)   & •	Rolling toy spindle drive system.\par•	Rolling toy deflector and method of use.\par•	Rolling toy anti-locking system and method.\par•	Rolling toy spinner reel.
\\
		\midrule
		Grinding \& Polishing\par(68$^{th }$ nearest) & 76.4\% (382/500)   & •	Fixture for rolling toy sleeves.\par•	Deep rolling toy arm with interchangeable rolling force.\par•	Nozzle device for the rolling of a rolling toy\par•	Children's rolling toy mill with disconnected work stations.
\\
		\midrule
		Fuels \& Lubricants\par(100$^{th }$ nearest) & 68.6\% (343/500)   & •	Lubricant for rolling toy made from colored aluminum alloy and coated with an acetylene-based additive.\par•	Toy diesel fuel production system and rolling toy vehicle.\par•	Toy with rolling bearing and friction mechanism.\par•	Electrical tool and planer for use in a rolling toy.
\\

		\bottomrule
	\end{tabular}
	\label{tab:table1}
\end{table}

Furthermore, as addressed by \citet{luo2019computer,luo2021guiding}, when ideating with external knowledge, the design ideas that synthesize far-field source knowledge could be more novel compared to those utilizing near-field stimuli. Therefore, our hypothesis is that the ideas generated by GPT-2 may follow the same pattern, i.e., far-domain virtual expert models may generate ideas with overall more novelty than near-domain virtual expert models.\par
To test the hypothesis, we employ TechNet \citep{sarica2020technet} as a tool to measure idea novelty. When measuring term-term relevancy in TechNet, the lower relevancy score means the given pair of terms are more semantically distant in the design context. Thus, by extracting all term pairs in a generated idea and calculating their semantic distances, the minimum value among them can be used to estimate the degree of novelty regarding the knowledge distance within the idea. The lower minimum score means a more novel idea. Figure 5 (a) reports the results of such evaluation.\par
\begin{figure}
	\centering
	\includegraphics[width = 12cm]{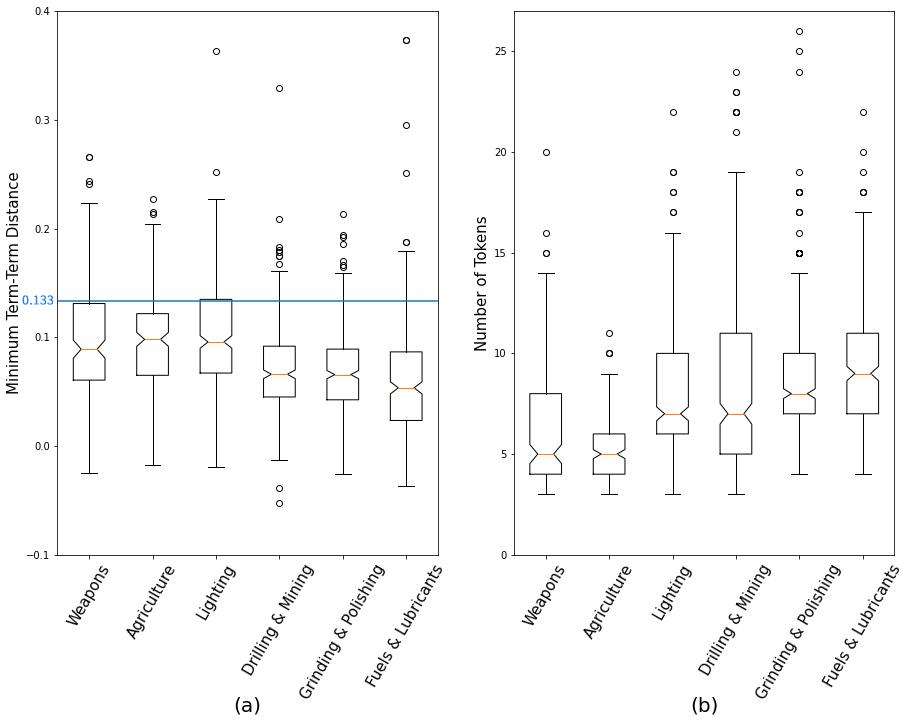}
	\caption{Distribution of the minimum term-term semantic distance (a) and number of tokens (b) of the generated ideas.}
	\label{fig:fig5}
\end{figure}

In TechNet, as reported by \citet{sarica2020technet}, the estimated mean value of the term relevancy score is 0.133 (shown as the blue line). It means the ideation using either near-field or far-field models achieves an overall good novelty. Note that this mean value was calculated using 108 different pairs of terms. In real contexts, the more relevant pairs will appear more commonly, thus making the mean value even higher. Moreover, according to Figure 5 (a), ideas generated by the far-field domain expert models achieve generally lower relevancy scores than those by near-field models, thus verifying our hypothesis. However, as shown in Figure 5 (b), far-field models tend to generate longer ideas which are likely to contain more details. The richer details could also be the source of the lower minimum relevancy because it could increase the probability of the co-occurrence of terms with low relevancy.\par
In Figure 6, we present four sketches drawn by the first author that are derived from the ideas generated from our near-field and far-field expert models, representing the initial embodiment designs of the concepts. Figure 6 (a) is a moving and rolling dartboard that provides a challenging game experience, it has two boards on the sides and can switch sides by turning around; (b) is a rolling light source that can move around with the target that it aims to lighten, and with a well-designed algorithm it will also be able to focus the light on the target while moving; (c) is a rolling reel robot that can lay cables or wires in any desired routes, which could be helpful to lay cables in narrow places that human cannot reach; (d) applies nozzles in rolling toys that could entirely change the way that they roll. This is done by using 8 nozzles that aim at different directions to create force in those directions, controlling the timing and strength of the nozzles will result in different movements. All the above concepts are selected and developed from the ideas generated by our method which are listed in Table 1. The initial generated ideas of the concepts are shown in the caption of Figure 6.\par
\begin{figure}
	\centering
	\includegraphics[width = 10cm]{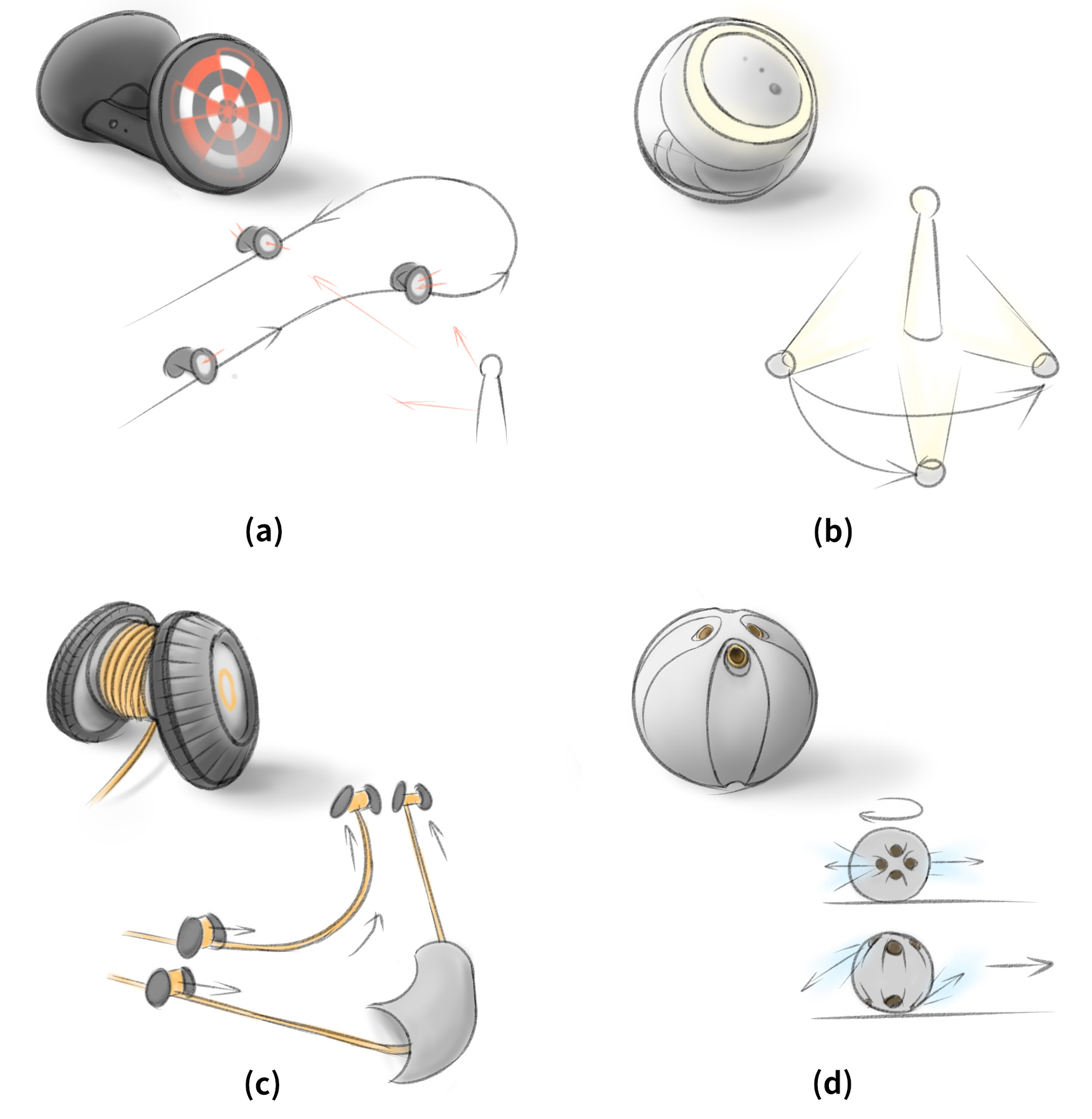}
	\caption{Sketches based on the generated ideas: (a) Rolling toy dart board capable of making turns (weapons); (b) Lighting device for rolling and adjusting a light source (lighting); (c) Rolling toy spinner reel (drilling \& mining); (d) Nozzle device for the rolling of a rolling toy (grinding \& polishing)}
	\label{fig:fig5}
\end{figure}

\section{Discussion and Future Work}
The proposed study explores a novel generative approach to design ideation which utilizes the fine-tuning mechanism of GPT to leverage knowledge understanding and domain synthesis. The method’s capability of generating novel and understandable ideas is verified in a rolling toy design project. We also show that the minimum term-term relevancy score can be used as an evaluation metric for idea novelty. In real ideation practice, not all output from our approach can be directly usable for progressing into the next stage of the design process. Designers may come across unfamiliar terms in the output that indicate unfamiliar knowledge, and therefore need further investigation to make sense of the ideas. The output may also vary in quality and some of the results could make less sense, especially for those generated with far-field knowledge. The proposed method is not expected to be a ‘machine designer’ that design concepts automatically, but a design assistant that ideate and brainstorm with human designers to boost their creativity. Thus, designers’ efforts are still required to develop sketches as in Figure 6 from generated ideas as in Table 1. \par
However, a statistical assessment approach regarding the performance of the idea generation and the quality of ideas is still a challenge in this context. Natural language understanding and sentence embedding techniques (e.g., BERT) could be an option for bridging this gap and will be explored in future works. Moreover, as an initial attempt at idea generation using natural language generation, this paper does not validate the choices for the subsystems and how they might affect the performance, e.g., what if we use GPT-3 to do few-shot learning for the generation of ideas? This will also be tested in the future with both machine and human calibration.\par
In sum, we hope this preliminary work, by using generative transformers to derive creative artificial intelligence for automatic design generation, open a new path of research for data-driven innovation \citep{luo2022data}.

\bibliographystyle{unsrtnat}
\bibliography{bibliography-bibtex.bib}  

\begin{thebibliography}{15}
\providecommand{\natexlab}[1]{#1}
\providecommand{\url}[1]{\texttt{#1}}
\expandafter\ifx\csname urlstyle\endcsname\relax
  \providecommand{\doi}[1]{doi: #1}\else
  \providecommand{\doi}{doi: \begingroup \urlstyle{rm}\Url}\fi

\bibitem[Viswanathan et~al.(2016)Viswanathan, Tomko, and
  Linsey]{viswanathan2016a}
Vimal Viswanathan, Megan Tomko, and Julie Linsey.
\newblock A study on the effects of example familiarity and modality on design
  fixation.
\newblock \emph{AI EDAM}, 30\penalty0 (2):\penalty0 171--184, 2016.

\bibitem[White et~al.(2012)White, Wood, and Jensen]{white2012from}
Christina~K White, Kristin~L Wood, and Dan Jensen.
\newblock From brainstorming to c-sketch to principles of historical
  innovators: ideation techniques to enhance student creativity.
\newblock \emph{Journal of STEM Education: Innovations and Research},
  13\penalty0 (5), 2012.

\bibitem[Yilmaz et~al.(2016)Yilmaz, Daly, Seifert, and
  Gonzalez]{yilmaz2016evidencebased}
Seda Yilmaz, Shanna~R Daly, Colleen~M Seifert, and Richard Gonzalez.
\newblock Evidence-based design heuristics for idea generation.
\newblock \emph{Design Studies}, 46:\penalty0 95--124, 2016.

\bibitem[Deldin and Schuknecht(2014)]{deldin2014the}
Jon-Michael Deldin and Megan Schuknecht.
\newblock The asknature database: enabling solutions in biomimetic design.
\newblock In \emph{Biologically inspired design}, pages 17--27. Springer, 2014.

\bibitem[Luo et~al.(2019)Luo, Sarica, and Wood]{luo2019computer}
Jianxi Luo, Serhad Sarica, and Kristin~L Wood.
\newblock Computer-aided design ideation using innogps.
\newblock In \emph{International Design Engineering Technical Conferences and
  Computers and Information in Engineering Conference}, volume 59186, page
  V02AT03A011. American Society of Mechanical Engineers, 2019.

\bibitem[Luo et~al.(2021)Luo, Sarica, and Wood]{luo2021guiding}
Jianxi Luo, Serhad Sarica, and Kristin~L Wood.
\newblock Guiding data-driven design ideation by knowledge distance.
\newblock \emph{Knowledge-Based Systems}, 218:\penalty0 106873, 2021.

\bibitem[Sarica et~al.(2020)Sarica, Luo, and Wood]{sarica2020technet}
Serhad Sarica, Jianxi Luo, and Kristin~L Wood.
\newblock Technet: Technology semantic network based on patent data.
\newblock \emph{Expert Systems with Applications}, 142:\penalty0 112995, 2020.

\bibitem[Sarica et~al.(2021)Sarica, Song, Luo, and Wood]{sarica2021idea}
Serhad Sarica, Binyang Song, Jianxi Luo, and Kristin~L Wood.
\newblock Idea generation with technology semantic network.
\newblock \emph{AI EDAM}, 35\penalty0 (3):\penalty0 265--283, 2021.

\bibitem[Chakrabarti et~al.(2011)Chakrabarti, Shea, Stone, Cagan, Campbell,
  Hernandez, and Wood]{chakrabarti2011computer}
Amaresh Chakrabarti, Kristina Shea, Robert Stone, Jonathan Cagan, Matthew
  Campbell, Noe~Vargas Hernandez, and Kristin~L Wood.
\newblock Computer-based design synthesis research: an overview.
\newblock \emph{Journal of Computing and Information Science in Engineering},
  11\penalty0 (2), 2011.

\bibitem[Regenwetter et~al.(2022)Regenwetter, Nobari, and
  Ahmed]{regenwetter2021}
Lyle Regenwetter, Amin~Heyrani Nobari, and Faez Ahmed.
\newblock Deep generative models in engineering design: A review.
\newblock \emph{Journal of Mechanical Design}, 144\penalty0 (7):\penalty0
  071704, 2022.

\bibitem[Duan et~al.(2020)Duan, Zhao, Zhou, Qiu, and Liu]{duan2020a}
Jiajia Duan, Hui Zhao, Qian Zhou, Meikang Qiu, and Meiqin Liu.
\newblock A study of pre-trained language models in natural language
  processing.
\newblock In \emph{2020 IEEE International Conference on Smart Cloud
  (SmartCloud)}, pages 116--121. IEEE, 2020.

\bibitem[Radford et~al.(2019)Radford, Wu, Child, Luan, Amodei, Sutskever,
  et~al.]{radford2019language}
Alec Radford, Jeffrey Wu, Rewon Child, David Luan, Dario Amodei, Ilya
  Sutskever, et~al.
\newblock Language models are unsupervised multitask learners.
\newblock \emph{OpenAI blog}, 1\penalty0 (8):\penalty0 9, 2019.

\bibitem[Brown et~al.(2020)Brown, Mann, Ryder, Subbiah, Kaplan, Dhariwal,
  Neelakantan, Shyam, Sastry, Askell, et~al.]{brown2020language}
Tom Brown, Benjamin Mann, Nick Ryder, Melanie Subbiah, Jared~D Kaplan, Prafulla
  Dhariwal, Arvind Neelakantan, Pranav Shyam, Girish Sastry, Amanda Askell,
  et~al.
\newblock Language models are few-shot learners.
\newblock \emph{Advances in neural information processing systems},
  33:\penalty0 1877--1901, 2020.

\bibitem[Sharma and Li(2019)]{sharma2019self}
Prafull Sharma and Yingbo Li.
\newblock Self-supervised contextual keyword and keyphrase retrieval with
  self-labelling.
\newblock 2019.

\bibitem[Luo(2022)]{luo2022data}
Jianxi Luo.
\newblock Data-driven innovation: What is it.
\newblock \emph{IEEE Transactions on Engineering Management}, 2022.
\newblock \doi{https://doi.org/10.1109/TEM.2022.3145231}.

\end{thebibliography}






\end{document}